\documentclass{article}

\PassOptionsToPackage{numbers, compress}{natbib}

\usepackage{todonotes}
\usepackage[preprint]{neurips_2021}

\usepackage[utf8]{inputenc} 
\usepackage[T1]{fontenc}    
\usepackage{hyperref}       
\usepackage{url}            
\usepackage{booktabs}       
\usepackage{amsfonts}       
\usepackage{nicefrac}       
\usepackage{microtype}      
\usepackage{xcolor}         

\usepackage{graphicx}
\usepackage{booktabs} 
\usepackage{microtype}
\usepackage{subcaption}
\usepackage{multirow}
\usepackage{tabularx}
\usepackage{xspace}

\title{Why is Pruning at Initialization Immune to Reinitializing and Shuffling?}

\author{%
  Sahib Singh\\
  ML Collective, Ford Motor Company\\
  \texttt{sahibsingh570@gmail.com} \\
  \And
  Rosanne Liu \\
  ML Collective \\
  \texttt{rosanne@mlcollective.org} \\
}

\newcommand{\wass}{\ensuremath{\mathsf{Wd}}\xspace}

\begin{document}

\maketitle
    
\begin{abstract}
Recent studies assessing the efficacy of pruning neural networks methods~\cite{frankle2020pruning, su2020sanity} uncovered a surprising finding: when conducting ablation studies on existing pruning-at-initialization methods, namely SNIP~\cite{lee2018snip}, GraSP~\cite{wang2020picking}, SynFlow~\cite{tanaka2020pruning}, and magnitude pruning, performances of these methods remain unchanged and sometimes even improve when randomly shuffling the mask positions within each layer (Layerwise Shuffling) or sampling new initial weight values (Reinit), while keeping pruning masks the same. We attempt to understand the reason behind such network immunity towards weight/mask modifications, by studying layer-wise statistics before and after randomization operations. We found that under each of the pruning-at-initialization methods, the distribution of unpruned weights changed minimally with randomization operations.


\end{abstract}

\section{Introduction}

Modern deep neural networks come with enormous compute and memory constraints which makes them really difficult to deploy on embedded systems \cite{han2015learning}. One approach for reducing such constraints is neural network pruning. Even though pruning has been studied since the 1980's \cite{PhysRevA.39.6600, mozer1989using, karnin1990simple}, we have seen a recent resurgence due to rise of deep learning \cite{blalock2020state}. While pruning can occur at various stages including \emph{before} training, \emph{during} training and \emph{after} training, in this work we focus on methods which prune before the training process begins, that is, pruning at initialization. We took a close look at what distributional properties are preserved or shifted before and after the pruning, with which we hope to uncover a recently reported intriguing observation where the pruning-at-initialization methods seem to be immune to weight re-shuffling and re-initializing.  

Recent works ~\cite{frankle2020pruning, su2020sanity} studying neural network pruning revealed a bewildering finding showcasing how state-of-the-art pruning-at-initialization methods perform equally well even after we a) randomly shuffle which weights they prune in each layer, or b) reinitialize the unpruned weights. Based on this they conclude that even though these pruning at initialization methods propose metrics to select specific weights to prune, their performance in fact doesn't degrade as long as they prune while maintaining the same layer-wise proportions as the unmodified pruning method.

The focus of our work is to understand why these ablations (Layerwise Shuffling, Reinit) of those methods don't seem to impact the method's efficacy. Since both ablations rely on a defined weight distribution, it is straightforward to consider calculating the distributional difference before and after these operations. Therefore, we compare the post-pruning distributions of weights of these ablations with the unmodified pruned model's weight distribution, for each of the pruning method. We use Wasserstein distance (\wass) as a measure of distributional similarity. The higher the \wass, the lower the similarity among distributions and vice-versa. To make sense of its value, we use the \wass{} of Random pruning with respect to unmodified pruning as a baseline for comparison.

\section{Background}
The work is built upon  pruning-at-initialization techniques, randomization treatments, and distributional distance measures. 
For the former two we stay close to what's used in recent literature~\cite{frankle2020pruning, su2020sanity}. As to similarity measures for weight distributions, we tried quite a few metrics including mutual information score, KL Divergence~\cite{Joyce2011}, Wasserstein Distance, L1 Norm and Total Variation. After some trial and error we decided to go with Wasserstein Distance since it proved to be the most robust to outliers.

\subsection{Pruning-at-initialization Methods}
\label{sec:prune}
Neural network pruning is a model compression technique aiming to reduce the size of a model's trainable parameters without too much degradation in performance.
Pruning-at-initialization methods, especially those covered in this paper and listed below, work by assigning each trainable parameter, $w$, a score, $z$, before any training step is performed. A pruning decision on $w$ is made according to the magnitude of $z$. We follow the mathematical notations used in~\cite{frankle2020pruning}, while omitting the layer index for simplification.

\emph{Magnitude}: This method issues each weight its magnitude $z = |w|$ as its score and removes those with the lowest scores.

\emph{SNIP} \cite{lee2018snip}: This method samples training data, computes gradients $g$ for each layer, issues scores $z = |g \bigodot w|$, and removes weights with the lowest scores in one iteration.

\emph{GraSP} \cite{wang2020picking}: This method samples training data, computes the Hessian-gradient product $h$ for each layer, issues scores $z = -w\bigodot h$, and removes weights with the highest scores in one iteration. Our work focuses on the absolute values of this method used denoted as \emph{GraspAbs} in \cite{frankle2020pruning}.

\emph{SynFlow} \cite{tanaka2020pruning}: This method replaces the weights $w$ with $|w|$. It forward propagates an input of $1$’s, computes the sum $R$ of the logits, and computes the gradients $r$ of $R$. It issues scores $z= |r \bigodot  w|$ and removes weights with the lowest scores.

All methods when noted "unmodified" are implemented as stated above, without the randomization treatment that is the focus of this study. Another pruning techniques used as a baseline in this paper is \emph{Random Pruning} which prunes each weight independently with a predefined probability for the whole network. 

\subsection{Randomization Treatments}
Randomization treatments to either the pruning mask or the underlying weights have been used to test the robustness of pruning methods. In~\cite{frankle2020pruning}, the authors used \emph{Randomly shuffling}, \emph{Reinitialization} and \emph{Inversion} at the ablations section. In another contemporary work~\cite{su2020sanity}, the authors adopted \emph{Layerwise rearrange} (shuffling all the weights/masks in a layer, in fact equivalent to \emph{Randomly shuffling} in~\cite{frankle2020pruning}) and \emph{Layerwise weights shuffling} (defined as shuffling only unmasked weights). In this paper we simplify the terminology to use \emph{Layerwise Shuffling} and \emph{Reinit}, as explained below.

\paragraph{Layerwise Shuffling} This is equivalent to the term \emph{Randomly shuffling} in~\cite{frankle2020pruning} and \emph{Layerwise rearrange} in~\cite{su2020sanity}. It refers to randomly shuffling the pruning mask within each layer. Basically, once the pruning mask has been obtained, such mask is randomly shuffled across every layer and this layerwise-shuffled network is then trained.
It's worth noting that unlike random pruning which basically rearranges the connections in the entire network, Layerwise Shuffling ensures the number of preserved weights in each layer stays the same as the unmodified pruned network \cite{su2020sanity}. To put things into context, if the original mask prunes 20\% weights in the first layer and 30\% in the succeeding layer then Layerwise Shuffling will also ensure those layerwise pruning ratios are maintained. 
Recent literature~\cite{frankle2020pruning, su2020sanity} proves that the accuracy of network post Layerwise Shuffling is unaffected compared to unmodified pruning hence the per-weight decisions made by the methods can be replaced by the per-layer fraction of weights it prunes.

\paragraph{Reinit} This is equivalent to the same term, \emph{Reinitialization} in~\cite{frankle2020pruning}. \emph{Reinit} involves sampling a new initialization for the pruned network from the same distribution as the original network. Reinit takes place after the initial pruning mask has been created based on the original weight values. The mask is then applied to the newly sampled (re-initialized) weights and the network is trained from then on. By conducting experiments using Reinit we can test if the networks produced by the given pruning methods are sensitive to the specific initial values of their weights. Earlier work by  \cite{frankle2020pruning} empirically demonstrates how all early pruning methods are unaffected by reinit since accuracy is the same irrespective of whether the network is trained with the original initialization or a newly sampled initialization.

\subsection{Wasserstein Distance}
Wasserstein distance (\wass) is a measure of the distance between two probability distributions. It can be understood as the minimum cost or effort required to transform one probability distribution to another. 
The wasserstein distance between two probability measures $\mu$ and $\nu$ is defined as:
\begin{equation}
    \wass(\mu,\nu) = \inf {\mathbb E}\left[ d(X,Y) \right]
\end{equation}
where $d$ is a metric, and $\mathbb {E}[\cdot]$ denotes the expected value of a random variable \cite{RubioWasserstein}. The infimum is taken over all joint distributions of the random variables X and Y with marginals $\mu$ and $\nu$ respectively.


\section{Experiments}

\begin{figure*}
    \centering
    \begin{subfigure}{0.49\textwidth}
     \centering
     \includegraphics[width=\textwidth]{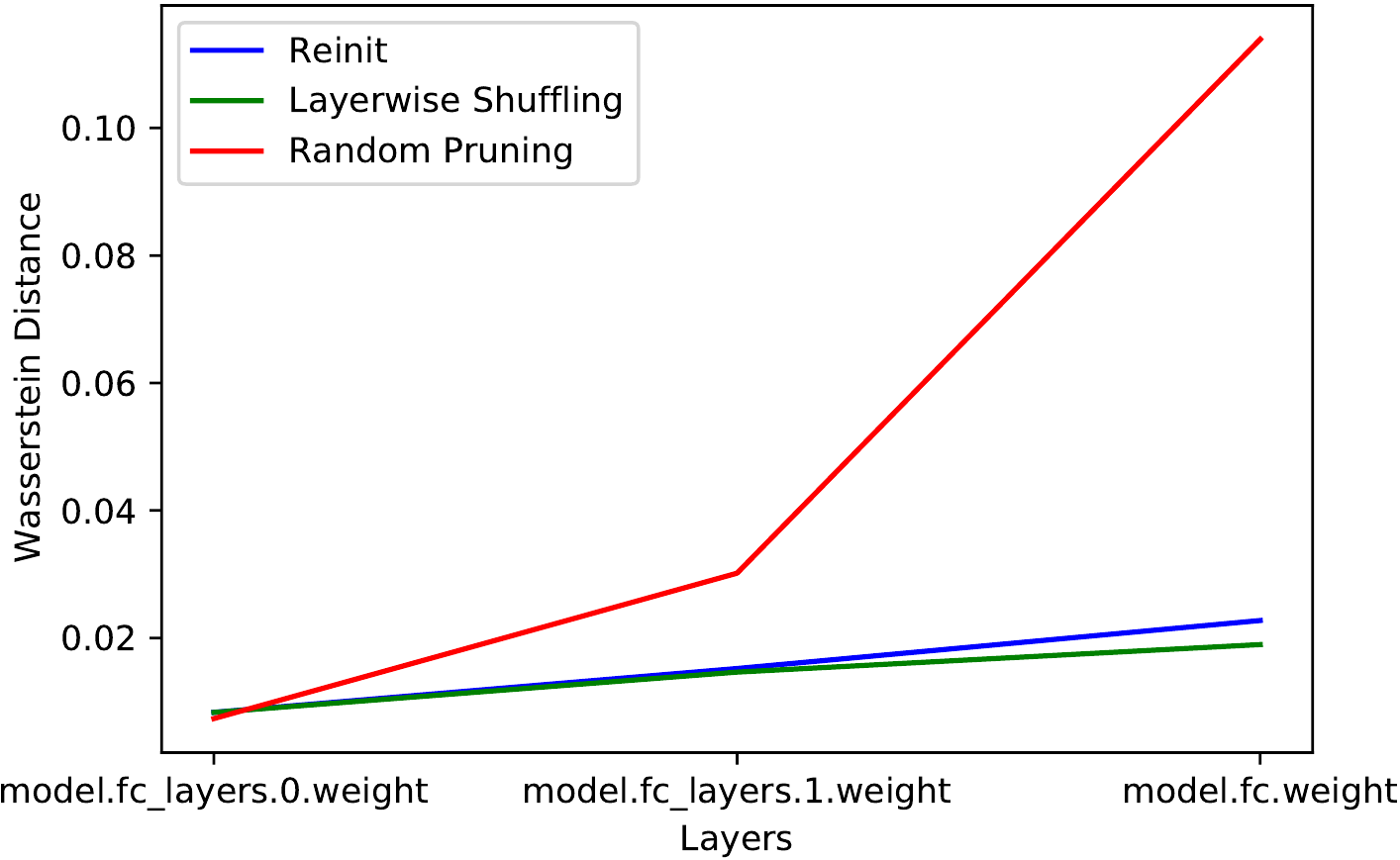}
     \caption{Magnitude Pruning}
     \label{fig:lenet_magnitude}
    \end{subfigure}
    \hfill
    \begin{subfigure}{0.49\textwidth}
     \centering
     \includegraphics[width=\textwidth]{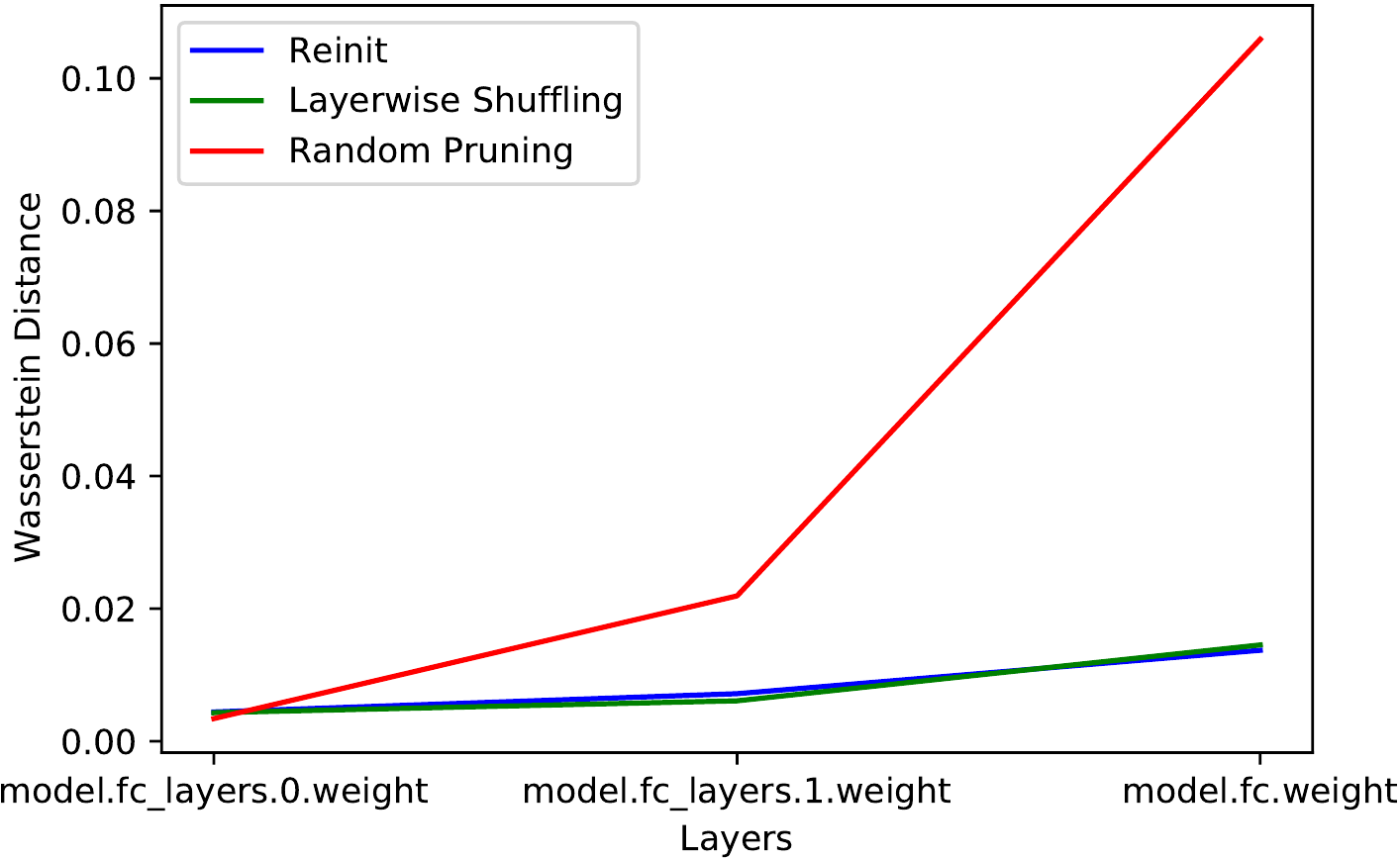}
     \caption{SNIP}
     \label{fig:lenet_snip}
    \end{subfigure}
    \begin{subfigure}{0.49\textwidth}
     \centering
     \includegraphics[width=\textwidth]{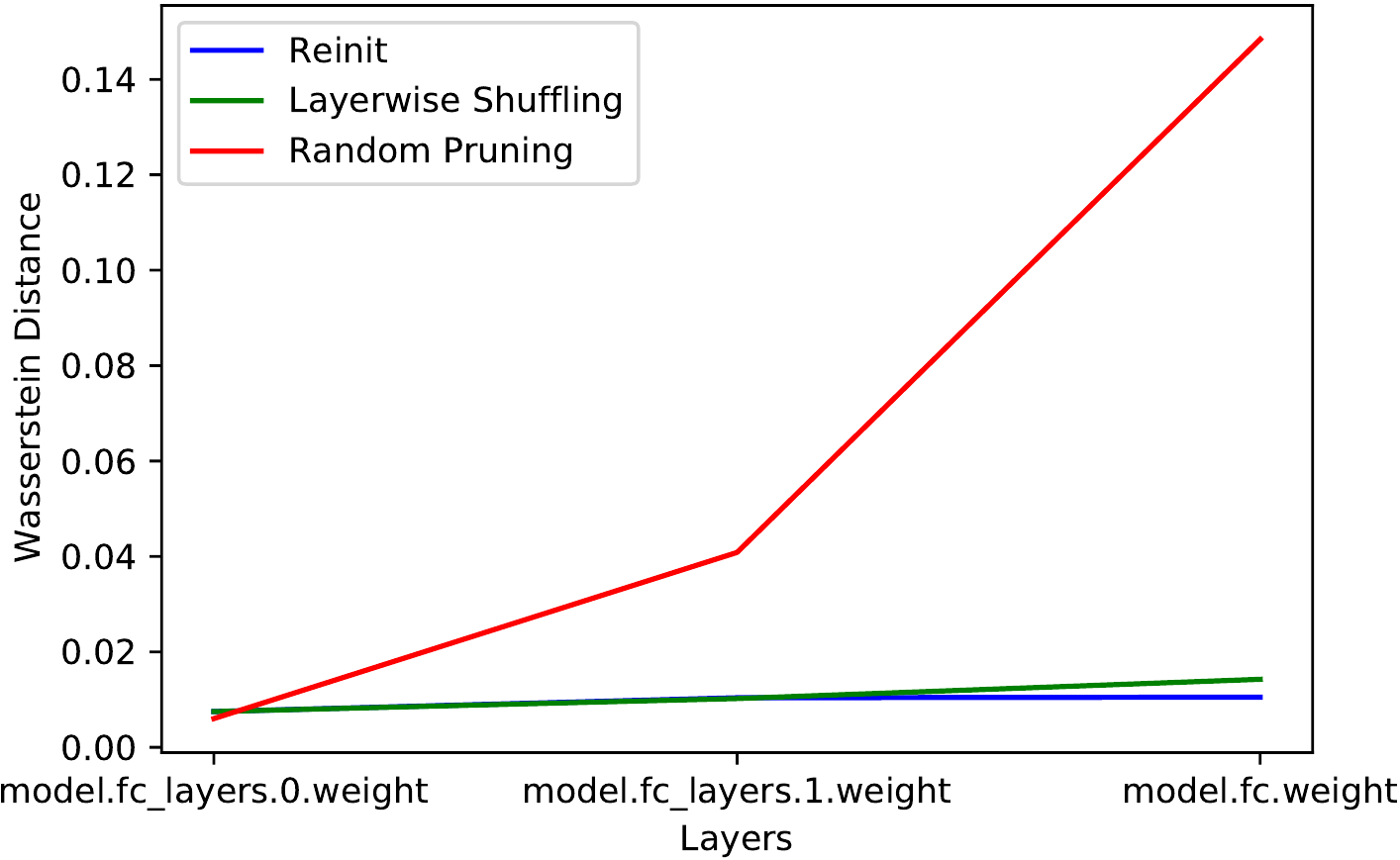}
     \caption{Synflow}
     \label{fig:lenet_synflow}
    \end{subfigure}
    \hfill
    \begin{subfigure}{0.49\textwidth}
     \centering
     \includegraphics[width=\textwidth]{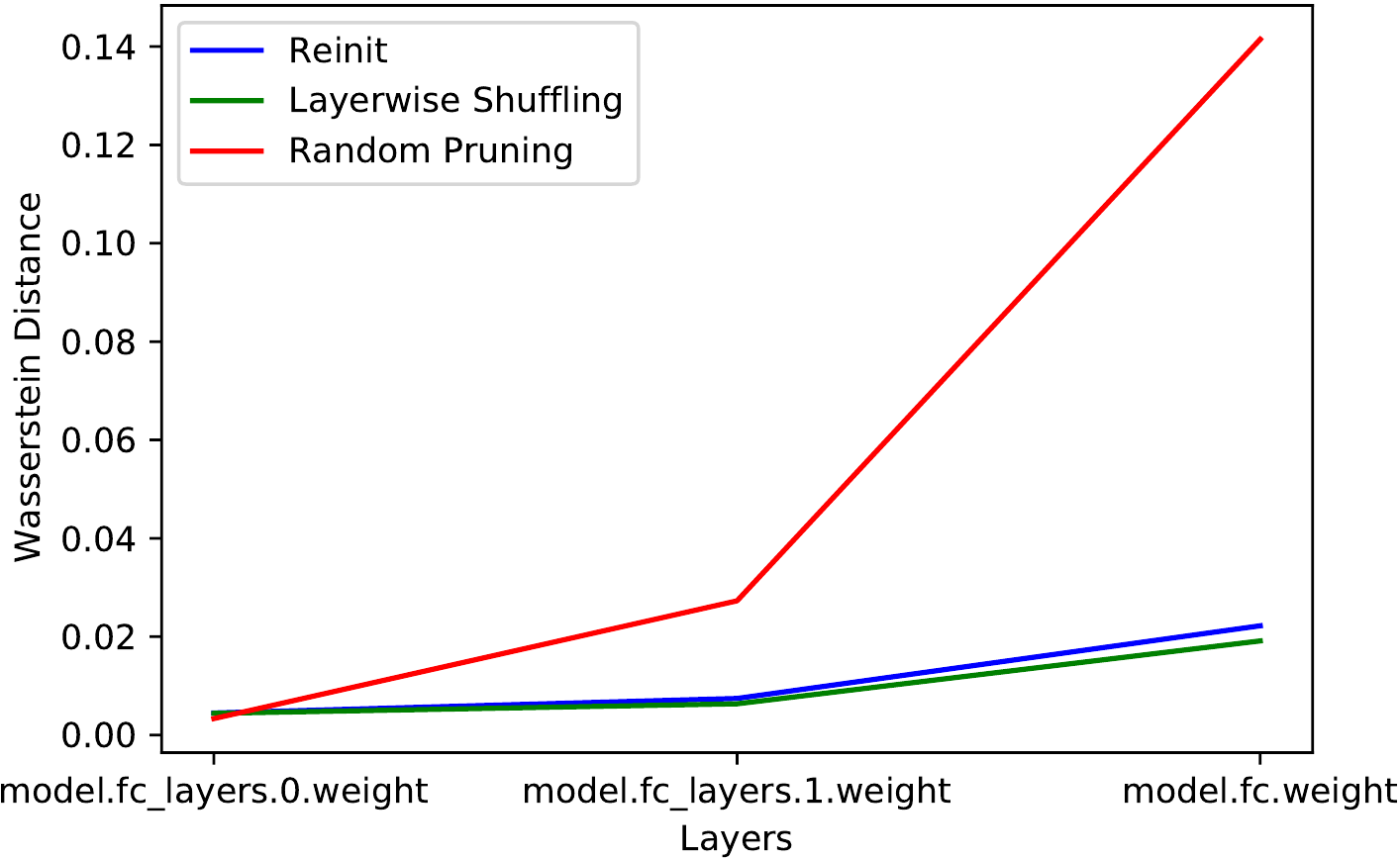}
     \caption{GraspAbs}
     \label{fig:lenet_graspabs}
    \end{subfigure}
    \caption{Layerwise \wass{} of LeNet trained on MNIST for all four pruning methods (a-d). Blue, green, red lines represent, respectively, \wass(Reinit, Unmodified), \wass(Layerwise Shuffling, Unmodified), and \wass(Random Pruning, Unmodified). Both Randomize and Reinit treatments modify weight distribution minimally, compared to random pruning.}
    \label{fig:lenet}
    
\end{figure*}

\begin{figure*}
    \centering
    \begin{subfigure}{0.49\textwidth}
     \centering
     \includegraphics[width=\textwidth]{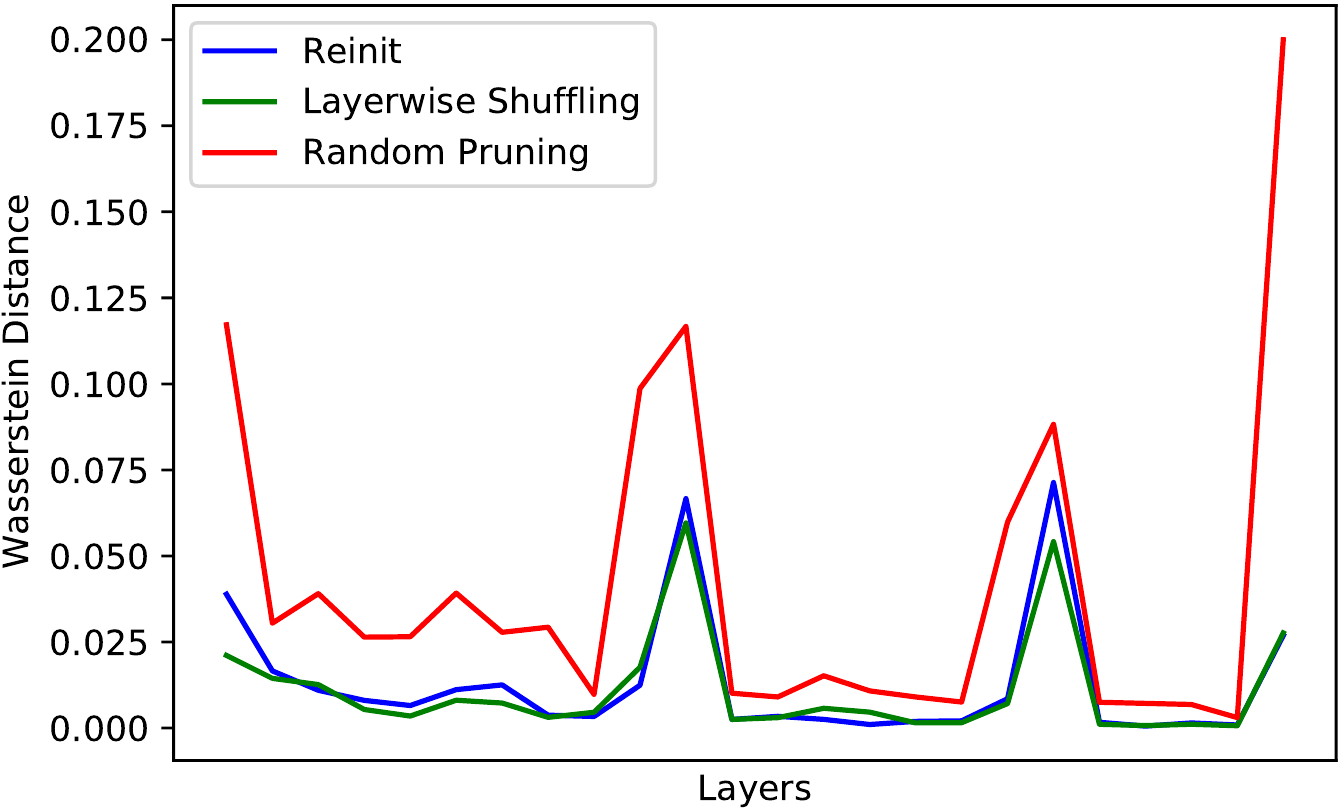}
     \caption{Magnitude Pruning}
     \label{fig:resnet_magnitude}
    \end{subfigure}
    \hfill
    \begin{subfigure}{0.49\textwidth}
     \centering
     \includegraphics[width=\textwidth]{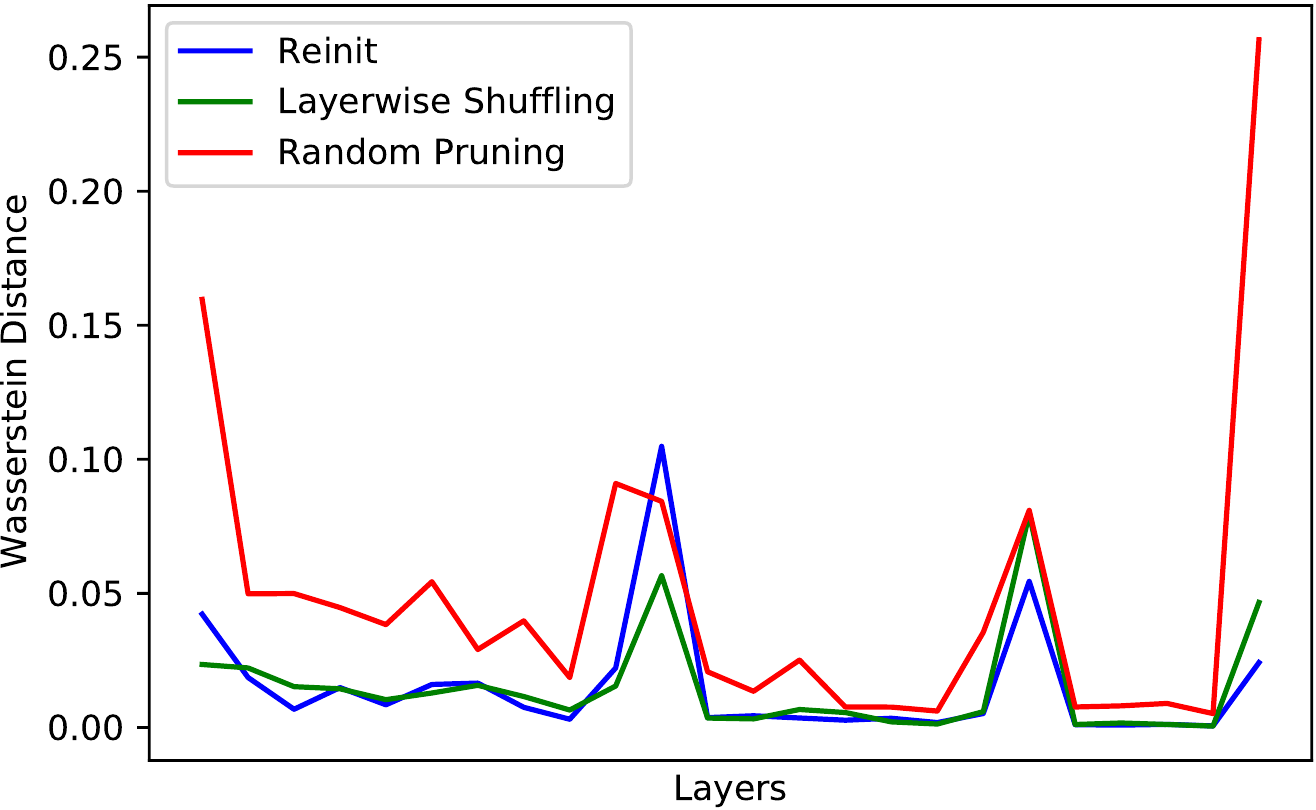}
     \caption{SNIP}
     \label{fig:resnet_snip}
    \end{subfigure}
    \begin{subfigure}{0.49\textwidth}
     \centering
     \includegraphics[width=\textwidth]{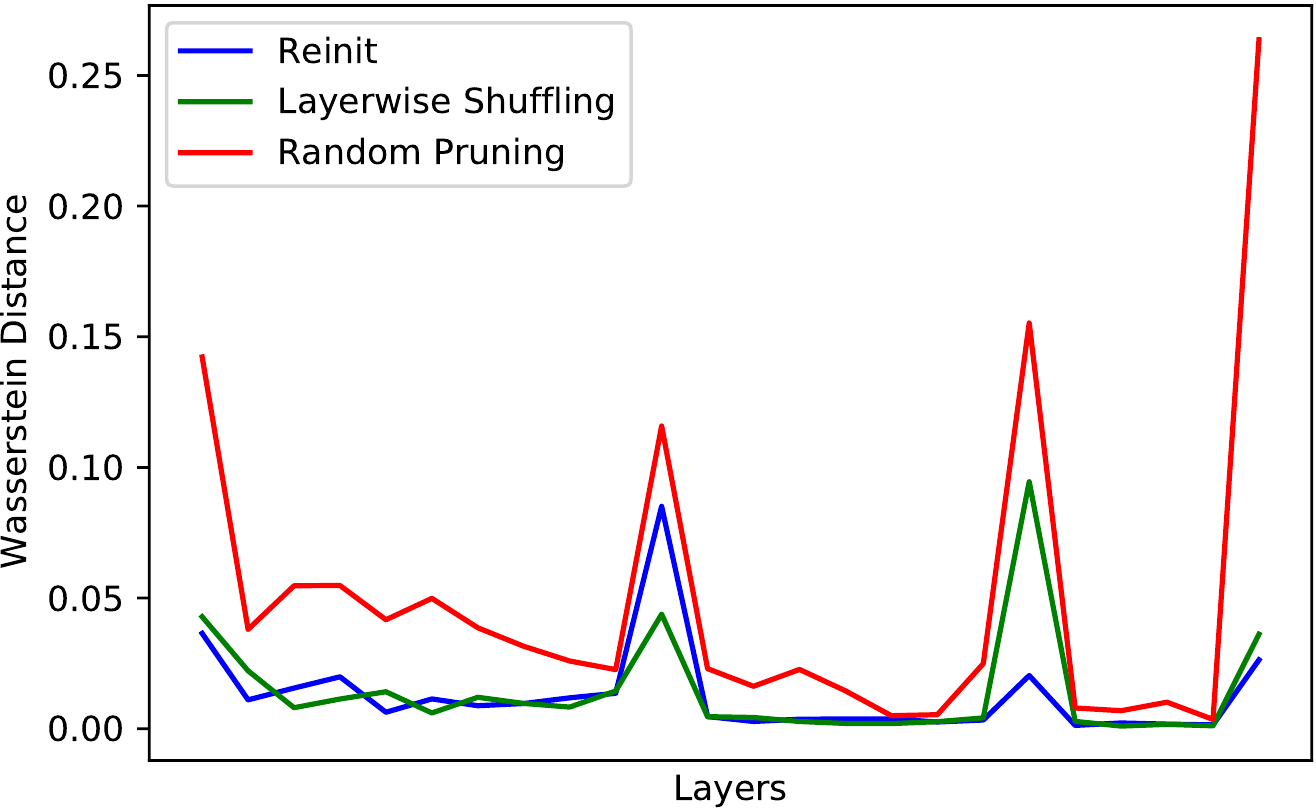}
     \caption{Synflow}
     \label{fig:resnet_synflow}
    \end{subfigure}
    \hfill
    \begin{subfigure}{0.49\textwidth}
     \centering
     \includegraphics[width=\textwidth]{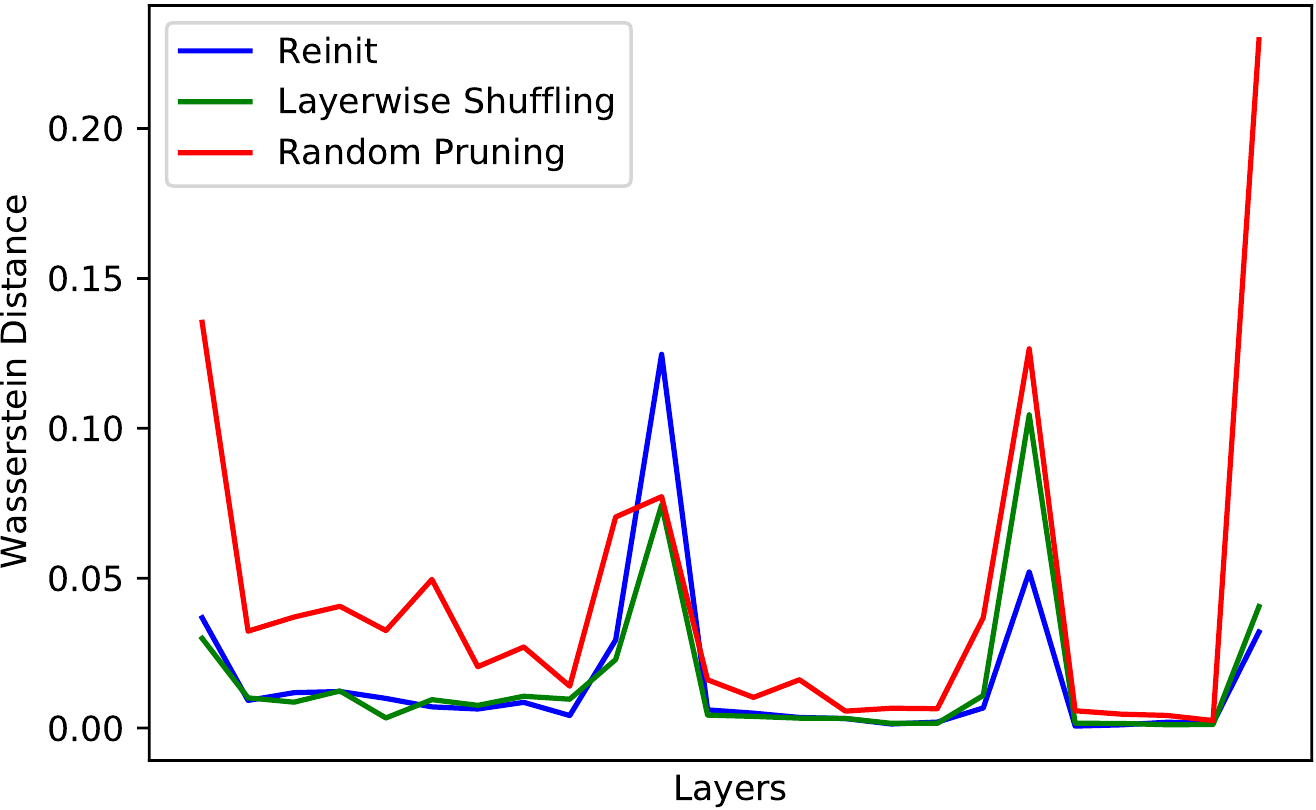}
     \caption{GraspAbs}
     \label{fig:resnet_graspabs}
    \end{subfigure}
    \caption{Layerwise distributional similarity of ResNet-20 trained on CIFAR for all four pruning methods (a-d). Blue, green, red lines represent, respectively, \wass(Reinit, Unmodified), \wass(Layerwise Shuffling, Unmodified), and \wass(Random Pruning, Unmodified). Both Randomize and Reinit treatments modify weight distribution minimally, compared to random pruning.}
    \vspace{-2mm}
    \label{fig:resnet}
    
\end{figure*}

\begin{table*}[th]
  \centering
  \caption{Average Wasserstein Distance (\wass{}) between each ablation treatment and its respective control (unmodified). The maximum \wass{} for each row is in bold. Results suggest that both treatments modify weight distribution minimally.}
  
  \begin{tabularx}{{\textwidth}}{@{}lrrrrr@{}}
  \toprule
   \textbf{Network} &\textbf{Pruning Methods}	&\textbf{Reinit}	&\textbf{Layerwise Shuffling}	&\textbf{Random Pruning}\\

    \midrule 
    \multirow{4}{*}{LeNet-MNIST}
    & Magnitude	&{0.0154} &{0.0139}	&{\textbf{0.0504}}\\ 
    & Snip &{0.0084} &{0.0083}	&{\textbf{0.0437}}\\ 
    & Synflow	&{0.0094} &{0.0106}	&{\textbf{0.0650}}\\ 
    & GraspAbs	&{0.0113} &{0.0099}	&{\textbf{0.0573}}\\ 
    \midrule 
    \multirow{4}{*}{ResNet-CIFAR}
    & Magnitude	&{0.0131} &{0.0111}	&{\textbf{0.0414}}\\ 
    & Snip &{0.0153} &{0.0151}	&{\textbf{0.0476}}\\ 
    & Synflow	&{0.0128} &{0.0146}	&{\textbf{0.0489}}\\ 
    & GraspAbs	&{0.0157} &{0.0157}	&{\textbf{0.0419}}\\   
    \midrule
  \end{tabularx}
  \label{tab:avgdist}
\end{table*}


Our experiments are built upon the code originally open-sourced by \cite{frankle2020pruning}. A subset of the original experiments are used in this paper, as stated below:
\begin{itemize}
\item Fully-connected LeNet-300-100 network with two hidden layers trained on MNIST.
\item ResNet-20 \cite{he2016deep} trained on CIFAR-10 dataset.
\end{itemize}
 
For each architecture and dataset combination, we run all four pruning methods as listed in Section~\ref{sec:prune}. For each method, as well as each randomization treatment (Layerwise Shuffling, Reinit), we plot the per-layer distribution similarity of weights with the treatment and without, measured by \wass{}. Results are shown in Figures~\ref{fig:lenet} and~\ref{fig:resnet}. We also add a baseline treatment, Random Pruning, to visibly show the difference and hence highlight the specialty of those ablation treatments under study.

A high \wass{} implies more effort to convert one distribution to the other, hence signalling lower similarity between the weight distributions under a treatment and that of the unmodified method. On the other hand, low \wass{} means high similarity, suggesting that the distribution of unpruned weights changed minimally.

As we can see from Figures~\ref{fig:lenet} and~\ref{fig:resnet}, Random Pruning consistently makes a much bigger shift in the distribution of weights compared to Reinit and Layerwise Shuffling: \wass(Random Pruning , Unmodified) $>>$ \{\wass(Reinit, Unmodified), \wass(Layerwise Shuffling, Unmodified)\}. 

Detailed \wass{} numbers, further averaged across all layers, for each pruning method under each treatment are shown in Table ~\ref{tab:avgdist}. We observe that this this holds true across all methods (Magnitude Pruning, SNIP, Synflow, GraspAbs) and all networks(LeNet-MNIST, ResNet-CIFAR). It is also worth noting that the blue and green lines representing \wass(Reinit, Unmodified) and \wass(Layerwise Shuffling, Unmodified), respectively, have highly comparable \wass{} scores.

\section{Conclusion}

This work is a follow-up on recent impactful research studying pruning methods~\cite{frankle2020pruning, su2020sanity}, attempting to answer the question the authors raised while performing an ablation study: why is it possible to reinitialize or
layerwise-shuffle the unpruned weights without hurting accuracy, for all pruning-at-initialization methods under study?

By re-running all pruning methods (Magnitude, Grasp, Snip, Synflow) included in~\cite{frankle2020pruning} on a subset of architectures and datasets (LeNet-MNIST, ResNet-CIFAR), we made observations on the weight distribution shift:

\begin{itemize}
    \item The variation in post-pruning distribution of weights for methods {Layerwise Shuffling, Reinit} with respect to unmodified pruning is significantly lower than the variation in distribution of weights between random pruning with respect to unmodified pruning (Table~\ref{tab:avgdist}). Since these methods have relatively less variation (compared to random pruning) with respect to the unmodified treatment, the performance of these models are also more likely to be similar to unmodified pruning in terms of accuracy.
    
    \item The distributions of weights between (Reinit, Unmodified) and (Layerwise Shuffling, Unmodified) are very similar as witnessed by the blue and green lines respectively in Figure~\ref{fig:lenet} and Figure~\ref{fig:resnet}. This reinforces how both of these methods are able to maintain similar performances to each other.  
\end{itemize}

To conclude, we attempt to understand why randomization treatments like Layerwise Shuffling and Reinit do not deteriorate the performance of pruning-at-initialization methods, and we look for answers from a distributional point of view. The initial finding is that the weight distribution after pruning, for each of the studied ablation treatment, is preserved, i.e. the remaining weight distribution under these randomization treatments is very similar to that of the unmodified pruning method. We do not claim that the similarity in weight distribution provides a full explanation, or that the distributional lens contains all the answers, but we believe it is a viable first step to uncover mysteries in pruning performances at large. Furthermore, our findings offer fresh insights which will be useful to keep in mind when designing new pruning or initialization algorithms.

\section*{Acknowledgements}
We thank Jonathan Frankle for providing generous assistance regarding his earlier work's code implementation. We thank reviewers of the Sparsity in Neural Networks Workshop 2021 for their valuable inputs and suggestions. We thank the ML Collective community for the ongoing support and feedback of this research.

\newpage
\bibliography{refs}
\bibliographystyle{unsrt}




\end{document}